\pdfoutput=1

\documentclass[preprint,5p,twocolumn,times]{elsarticle}

\usepackage{graphicx}
\usepackage{amssymb}

\usepackage{hyperref}
\usepackage[frozencache,draft=true]{minted}
\usepackage{breakurl}

\usepackage{lineno}

\journal{SSRN - Version 1.01}

\begin{document}
\sloppy
\begin{frontmatter}


\title{LexNLP: Natural language processing and information extraction\\ for legal and regulatory texts}



\author{Michael J Bommarito II}
\ead{mike@lexpredict.com}
\author{Daniel Martin Katz}
\ead{dan@lexpredict.com}
\author{Eric M Detterman}
\ead{eric@lexpredict.com}
\address{LexPredict, LLC}

\begin{abstract}
LexNLP is an open source Python package focused on natural language processing and machine learning for legal and regulatory text.  The package includes functionality to (i) segment documents, (ii) identify key text such as titles and section headings, (iii) extract over eighteen types of structured information like distances and dates, (iv) extract named entities such as companies and geopolitical entities, (v) transform text into features for model training, and (vi) build unsupervised and supervised models such as word embedding or tagging models.  LexNLP includes pre-trained models based on thousands of unit tests drawn from real documents available from the SEC EDGAR database as well as various judicial and regulatory proceedings.  LexNLP is designed for use in both academic research and industrial applications, and is distributed at \url{https://github.com/LexPredict/lexpredict-lexnlp}.
\end{abstract}


\end{frontmatter}


\section{Introduction}
\label{S:introduction}
Over the last two decades, many high-quality, open source packages for natural language processing and machine learning have been released.  Researchers and developers can quickly write applications in languages such as Java, Python, and R that stand on the shoulders of comprehensive, well-tested libraries like Stanford NLP (\cite{manning-EtAl:2014:P14-5}), OpenNLP (\cite{OpenNLP}), NLTK (\cite{BirdKleinLoper09}), spaCy (\cite{spacy2}), scikit-learn (\cite{sklearn_api}, \cite{scikit-learn}), Weka (\cite{hall09:_weka_data_minin_softw}), and gensim (\cite{rehurek_lrec}).  Consequently, for most domains, the rate of research has increased and the cost of application development has decreased.  

For some specialized areas like medicine and marketing, there are focused libraries and organizations like the BioMedICUS (\cite{biomedicus}), RadLex (\cite{langlotz2006radlex}), and the Open Health Natural Language Processing Consortium.  Law, however, has received substantially less attention than others, despite its ubiquity, societal importance, and specialized form.  LexNLP is designed to fill this gap by providing both tools and data for researchers and developers to work with real legal and regulatory text, including statutes, regulations, court opinions, briefs, contracts, and other legal work product.  

Law is a domain driven by language, logic, and conceptual relationships, ripe for computation and analysis (\cite{ruhl2017harnessing}).  However, in our experience, natural language processing and machine learning have not been applied as widely or fruitfully in legal as one might hope.  We believe that a key impediment to academic and commercial application has been the lack of tools that allow users to turn real, unstructured legal documents into structured data objects.  The goal of LexNLP is to make this task simpler, whether for the analysis of statutes, regulations, court opinions, briefs or the migration of legacy contracts to smart contract or distributed ledger systems.


\subsection{History}
LexNLP and its sister data repository, the LexPredict Legal Dataset, have been developed since 2015 by LexPredict, a legal technology and services company.  LexPredict open sourced both LexNLP, the Legal Dataset, and ContraxSuite, a document and contract analytics platform based on LexNLP, in 2017.  These repositories have all been been updated on a monthly release schedule on GitHub at \url{https://github.com/LexPredict/lexpredict-lexnlp}, and public documentation has been available at ReadTheDocs since release 0.1.6.  

\subsection{License and Support}
LexNLP has been developed thus far under an Affero GPL license to allow for maximum inclusion of open source software such as ghostscript, gensim-simserver, or spaCy.  However, just as spaCy has liberalized its license terms over time from AGPL with commercial release to an open MIT license, we have considered whether and if so when to convert to MIT or Apache licensing.  Support is provided through GitHub issue tracking and by email.

The Legal Dataset repository is distributed under the Creative Commons Attribution Share Alike 4.0 (CC-BY-SA 4.0), since many sources of data in this repository are retrieved or curated from Wikipedia.

\section{Design}
\label{S:design}
LexNLP is designed to provide a single, standardized Python interface for working with legal and regulatory text. However, this is not accomplished by rewriting all core linguistic and statistical methods from scratch.  Instead, LexNLP is designed to rely wherever possible on a small set of key libraries, including NLTK, scikit-learn, scipy.  These libraries have been written, documented, and tested by thousands of developers, and feature compatible licensing and stable APIs.

In selecting these packages, we were guided by the principles below:

\begin{enumerate}
\item \textbf{Standard open source license}: We strongly prefer packages with standard open source licensing options like MIT, Apache, or GPL-family licenses.
\item \textbf{High level of maturity}: We strongly prefer packages with mature code bases, including years of development and testing.
\item \textbf{High level of documentation}: We strongly prefer packages with well-documented code bases.
\item \textbf{Broad platform and hardware support}: We strongly prefer packages that are platform- and hardware-agnostic, avoiding operating system, CPU/GPU, or memory constraints.
\item \textbf{Broad language and character support}: We strongly prefer packages that natively support non-English as well as English text.
\item \textbf{Strong ecosystem}: We strongly prefer packages with large and active communities of developers and users.
\end{enumerate}

Below, we discuss key package selections.

\subsection{Natural Language Processing}
We selected NLTK for core natural language processing functionality.  NLTK is a mature Python project with over seventeen years of development, a large community of developers and users, detailed documentation, supplementary models and corpora, and an Apache License model for it code.  While some packages such as TreeTagger or Stanford NLP may offer better performance or wider language support for some tasks, their restricted source or license models make them difficult to redistribute.  Given its ubiquitous usage in research, we developed an optional interface to the Stanford NLP library, including POS and NER models; however, these are disabled by default and must be explicitly enabled at runtime.  

It should also be noted that the spaCy project has quickly grown to be a compelling alternative to NLTK.  While we are excited about the project's future, we judged its current maturity level, licensing strategy, and community support to be less stable than NLTK.  However, we have engineered our usage of NLTK to allow for its potential replacement, and will continue to monitor the development of spaCy.

\subsection{Machine Learning}
We selected scikit-learn for our core machine learning functionality.  Scikit-learn is another mature Python project, with over eleven years of development, a large community of developers and users including sponsoring corporations, and a permissive BSD license.  Scikit-learn is built on top of and interoperates with \textit{numpy} and \textit{scipy}, providing functionality for feature transformation, feature and target preprocessing, unsupervised, semi-supervised, and supervised modeling, and model selection and assessment.  Scikit-learn's largest deficiency relative to other machine learning packages is its minimal support for sophisticated ``deep learning'' models; however, this gap is due to scikit-learn's lack of GPU support, which aligns with our principle of broad platform and hardware support.  Further, as scikit-learn's FAQ states, we agree that ``much larger gains in speed can often be achieved by a careful choice of algorithms'' than by use of GPU.  ``Deep'' NLP research is proceeding at a rapid pace, however, and we have begun to evaluate optional support for the \textit{keras} package to enable GPU computation and ``deep'' models (\cite{chollet2015keras}).

\subsection{Language Support}
LexNLP is designed to support multiple languages and character sets across its feature set.  Currently, English language support is available and full German language support is targeted for release in 2018 with additional languages such as French, Spanish and Mandarin to be added thereafter. 

\subsection{Unit Testing and Code Coverage}
LexNLP is developed using continuous integration (CI) practices, including unit testing, code coverage analysis, and code style analysis.  Thousands of test records are available on GitHub in CSV format and results are verified with every commit.  Coding style is based on PEP8 and enforced through CI as well.

\section{LexNLP Package}
\label{S:package}

\subsection{Natural Language Processing}
LexNLP provides the following natural language processing capabilities and resources:

\begin{itemize}
\item \textbf{Stopwords}: As a specialized domain of communication, legal text features a number of high-frequency ``stopwords'' that do not occur commonly in English otherwise.  LexNLP currently distributes and uses legal stopwords based on analysis of hundreds of thousands of contracts from the US Securities and Exchange Commission's (SEC) EDGAR database.

\item \textbf{Collocations}: As with stopwords, collocations in legal text differ from those calculated from general English text.  LexNLP currently distributes and uses collocations based on analysis of hundreds of thousands of contracts from the SEC EDGAR database.  These collocations include six sets - the top 100, 1,000, and 10,000 bigram and trigrams.

\item \textbf{Segmentation}: LexNLP provides segmentation capabilities for documents, pages, paragraphs, sections, and sentences.  Document and section segmentation are provided through the identification of titles or headings; for example, locating text such as ``EMPLOYMENT AGREEMENT,'' ``APPENDIX A,'' or ``VII. Indemnification and Insurance.''  Paragraph and sentence segmentation are provided through Punkt models trained on hundreds of thousands of contracts from the SEC EDGAR database (\cite{Kiss_2006)unsupervised}).  Additionally, all segmentation models are fully customizable, and the sentence and title models can be retrained with a single method call.  LexNLP can optionally call Stanford NLP functionality such as the StanfordTokenizer, although this must be explicitly enabled at runtime.

\item \textbf{Tokens, Stems, and Lemmas}: LexNLP provides tokenization, stemming, and lemmatization capabilities through NLTK by default.  Tokens, stems, and lemmas can all be extracted from text, either as materialized lists or Python generators.  All methods support standard transforms including lowercasing and stopwording.  By default, these methods in English use the Treebank tokenizer, Snowball stemmer, and WordNet lemmatizer.  LexNLP can also optionally call Stanford NLP functionality such as the StanfordTokenizer, although this must be explicitly enabled at runtime.

\item \textbf{Parts of Speech}: LexNLP provides part of speech (PoS) tagging and extraction, including methods to locate nouns, verbs, adjectives, and adverbs.  All methods support standard transforms including lowercasing and lemmatizing.  LexNLP can also call Stanford NLP for StanfordPOSTagger, although this must be explicitly enabled at runtime.  We are currently annotating a large sample of documents and intend to release a legal-specific PoS tagging model for English and German in 2018.

\item \textbf{Character sequence and n-gram/skipgram Distributions}: LexNLP provides functionality to quickly generate character sequence distributions, token n-gram distributions, and token skipgram distributions.  These distributions can be used in the customization or development of models such as LexNLP's segmentation models or more sophisticated classification models.
\end{itemize}

\subsection{Information Extraction}
LexNLP provides the following information extraction capabilities and resources:

\begin{itemize}
\item \textbf{Addresses}: LexNLP provides a custom tag-based model for the extraction of addresses like ``2702 LOVE FIELD DR,'' including common street and building abbreviation disambiguation.

\item \textbf{Amounts}: LexNLP provides functionality for the extraction of non-monetary amounts such as ``THIRTY-SIX THOUSAND TWO-HUNDRED SIXTY-SIX AND 2/100'' or ``2.035 billion tons.''

\item \textbf{Citations}: LexNLP provides functionality for the extraction of common legal citations, such as ``10 U.S. 100'' or ``1998 S. Ct. 1.''  This functionality is based on data provided by the Free Law Project's Reporters Database (\cite{reportersdb}).

\item \textbf{Conditional statements}: LexNLP provides functionality for the extraction of conditional statements such as ``subject to ...'' or ``unless and until ....''  The full list of conditional statements in English includes:

\begin{itemize}
\item if [not]
\item when [not]
\item where [not]
\item unless [not]
\item until [not]
\item unless and until
\item as soon as [not]
\item provided that [not]
\item \ [not] subject to
\item upon the occurrence
\item subject to
\item conditioned on
\item conditioned upon
\end{itemize}

\item \textbf{Constraints}: LexNLP provides functionality for the extraction of constraint statements such as ``at most'' and ``no less than.''  The full list of constraint statements in English includes:

\begin{itemize}
\item after
\item at least
\item at most
\item before
\item equal to
\item exactly
\item first of
\item greater [of, than, than or equal to]
\item greatest [of, among]
\item smallest [of, among]
\item last of
\item least of
\item lesser [of, than, than or equal to]
\item \ [no] less [than, than or equal to]
\item maximum [of]
\item minimum [of]
\item more than [or equal to]
\item \ [no] earlier than
\item \ [no] later than
\item \ [not] equal to
\item \ [not] to exceed
\item within
\item exceed[s]
\item prior to
\item highest [of]
\item lowest [of]
\end{itemize}

\item \textbf{Copyrights}: LexNLP supports the extraction of copyrights such as ``(C) Copyright 2000 Acme''.

\item \textbf{Courts}: LexNLP supports the extraction of court references such as ``Supreme Court of New York'' or ``S.D.N.Y.''  This functionality relies on the inclusion of data from the LexPredict Legal Dataset, available at GitHub at \url{https://github.com/LexPredict/lexpredict-legal-dictionary}, and covers courts across the US, Canada, Australia, and Germany.  

\item \textbf{Dates}: LexNLP supports the extraction of date references such as ``June 1, 2017'' or ``2018-01-01.''  This functionality is provided through two approaches.  First, a forked and improved version of the datefinder package, available at \url{https://github.com/LexPredict/datefinder}, provides complex regular expression matching for common date formats.  This approach, \textsc{get\_raw\_dates}, is calibrated towards higher recall, resulting in many false positive records.  A second approach, \textsc{get\_dates}, uses a character distribution machine learning model to remove potential false positives from the result set, producing a higher quality result at the cost of runtime.

\item \textbf{Definitions}: LexNLP supports the extraction of definitions such as ``... shall mean ...'' and ``... is defined as ...''

\item \textbf{Distances}: LexNLP supports the extraction of distances such as ``fifteen miles'' or ``30.5 km.''

\item \textbf{Durations}: LexNLP supports the extraction of durations such as ``ten years'' or ``120 seconds.''  

\item \textbf{Geopolitical Entities}: LexNLP supports the extraction of geopolitical entities such as ``New York'' or ``Norway.''  This functionality relies on the inclusion of data from the LexPredict Legal Dataset, available at GitHub at \url{https://github.com/LexPredict/lexpredict-legal-dictionary}, and covers countries, states, and provinces in English, French, German, Spanish.

\item \textbf{Money and Currencies}: LexNLP provides functionality for the extraction of monetary amounts such as ``\$5'' or ``ten Euros.''  By default, only the following ISO 4217 currency codes and their corresponding Unicode symbols are extracted: USD, EUR, GBP, JPY, CNY/RMB, and INR.

\item \textbf{Percents and Rates}: LexNLP supports the extraction of percents or rates such as ``10.5\%'' and ``50 bps.''

\item \textbf{Personally Identifying Information (PII)}: LexNLP supports the extraction of PII such as phone numbers, addresses, and social security numbers.

\item \textbf{Ratios}: LexNLP supports the extraction of ratios or odds such as ``3:1'' or ``four to three.''

\item \textbf{Regulations}: LexNLP supports the extraction of regulatory references such as ``32 CFR 170'' or ``Pub. L. 555-666.''  This functionality relies partially on the inclusion of data from the LexPredict Legal Dataset, available at GitHub at \url{https://github.com/LexPredict/lexpredict-legal-dictionary}, for the identification of US state citations such as ``Mo. Rev. Stat.''

\item \textbf{Trademarks}: LexNLP supports the extraction of trademark references such as ``Widget(R)'' or ``Hal (TM).''

\item \textbf{URLs}: LexNLP supports the extraction of URL references such as ``www.acme.com/terms.''
\end{itemize}

\subsection{Word Embeddings and Text Classifiers}
LexNLP provides the following word embedding and text classifier capabilities and resources:

\begin{itemize}
\item \textbf{word2vec legal models}: LexNLP has been used to produce large word2vec models from SEC EDGAR material (\cite{DBLP:journals/corr/abs-1301-3781}), and these models are distributed through the Legal Dataset repository referenced above.  These CBOW models are all trained with gensim; vector sizes of 50, 100, 200, and 300 are distributed with windows of 5, 10, and 20.  These models have been available since October 2017.

\item \textbf{word2vec contract models}: In addition to word2vec models trained on broad text examples, some models are also trained on specific contract types.  gensim CBOW models with vector size 200 and window 10 are trained and distributed for samples of credit, employment, services/consulting, and underwriting agreements.  These models have been available since October 2017.

\item \textbf{doc2vec contract models}: LexNLP has been used to produce large doc2vec models from SEC EDGAR material (\cite{DBLP:journals/corr/LeM14}).  These models are scheduled for release and distribution in the 0.1.10 release, along with a forthcoming academic article.

\item \textbf{doc2vec opinion models}: LexNLP has been used to produce large doc2vec models from Federal and State court opinions.  These models are scheduled for release and distribution in the 0.1.10 release, along with a forthcoming academic article.

\item \textbf{Contract/non-contract classifier}: LexNLP and its doc2vec models have been used to train classifiers capable of classifying documents as either contracts or non-contracts.  These models are scheduled for release and distribution in the 0.1.11 release, along with a forthcoming academic article.

\item \textbf{Contract type classifier}: LexNLP and its doc2vec models have been used to train classifiers capable of classifying contracts among broad types such as service agreements, confidentiality agreements, or labor and employment agreements.  These models are scheduled for release and distribution in the 0.1.11 release, along with a forthcoming academic article.

\item \textbf{Clause classifier}: LexNLP and its word2vec models have been used to train classifiers capable of classifying clauses among broad types such as confidentiality, insurance, or assignment.  These models are scheduled for release and distribution in the 0.1.11 release, along with a forthcoming academic article.
\end{itemize}

\subsection{Lexicons and Other Data}
In additional to word embeddings, pre-trained classifiers, and geopolitical entities, the Legal Dataset repository also includes a range of other important resources, including:
\begin{itemize}
\item \textbf{Accounting Lexicon}: US GAAP, UK GAAP, US GASB, US FASB, and IFRS FASB.
\item \textbf{Financial}: Common English financial terms and aliases, e.g., ``American Depository Receipt'' and ``ADR''
\item \textbf{Geopolitical Actors and Bodies}: US Federal and State regulators and agencies, UK regulators and agencies, Australian regulators and agencies, Canadian regulators and agencies
\item \textbf{Legal Lexicon}: Common Law based on Black's Law Dictionary (1910 edition), top 1,000 common law terms based on English contracts from SEC EDGAR database, and terms from US state and federal codes
\item \textbf{Scientific}: Chemical elements, common compounds, and hazardous waste in multiple languages
\end{itemize}

While these resources are not required to use LexNLP generally, they can substantially improve the quality of research or applications.  

\section{Example Usage}
\label{S:examples}
The tools in LexNLP can be combined and deployed to solve a range of complex informatics problems. To demonstrate LexNLP functionality and API, we provide a simple example of usage on a Purchase and Sale Agreement retrieved from the SEC EDGAR database at \url{https://www.sec.gov/Archives/edgar/data/1469822/000119312513041312/d447090dex1052.htm}.  

Example text is presented below:

\textbf{Example 1}
\begin{quote}
THIS PURCHASE AND SALE AGREEMENT (this “Agreement”) is made to be effective as of October 12, 2012 (the “Effective Date”), by and between WESLEY VILLAGE DEVELOPMENT, LP, a Delaware limited partnership (“Seller”), and KBS-LEGACY APARTMENT COMMUNITY REIT VENTURE, LLC, a Delaware limited liability company (“Buyer”).
\end{quote}

\textbf{Example 2}
\begin{quote}
“Deposit” shall mean One Million Two Hundred Fifty Thousand and No/100 Dollars (\$1,250,000.00), consisting of, collectively, the first deposit of Two Hundred Fifty Thousand and No/100 Dollars (\$250,000.00) (the “First Deposit”), and the second deposit of One Million and No/100 Dollars (\$1,000,000.00) (the “Second Deposit”), to the extent Buyer deposits the same in accordance with the terms of Section 2.1, together with any interest earned thereon.
\end{quote}

\textbf{Example 3}
\begin{quote}
4.2.2    Release. By accepting the Deed and closing the Transaction, Buyer, on behalf of itself and its successors and assigns, shall thereby release each of the Seller Parties from, and waive any and all Liabilities against each of the Seller Parties for, attributable to, or in connection with the Property, whether arising or accruing before, on or after the Closing and whether attributable to events or circumstances which arise or occur before, on or after the Closing, including, without limitation, the following: (a) any and all statements or opinions heretofore or hereafter made, or information furnished, by any Seller Parties to any Buyer’s Representatives; and (b) any and all Liabilities with respect to the structural, physical, or environmental condition of the Property, including, without limitation, all Liabilities relating to the release, presence, discovery or removal of any hazardous or regulated substance, chemical, waste or material that may be located in, at, about or under the Property, or connected with or arising out of any and all claims or causes of action based upon CERCLA (Comprehensive Environmental Response, Compensation, and Liability Act of 1980, 42 U.S.C. §§9601 et seq., as amended by SARA (Superfund Amendment and Reauthorization Act of 1986) and as may be further amended from time to time), the Resource Conservation and Recovery Act of 1976, 42 U.S.C. §§6901 et seq., or any related claims or causes of action (collectively, “Environmental Liabilities”); and (c) any implied or statutory warranties or guaranties of fitness, merchantability or any other statutory or implied warranty or guaranty of any kind or nature regarding or relating to any portion of the Property. Notwithstanding the foregoing, the foregoing release and waiver is not intended and shall not be construed as affecting or impairing any rights or remedies that Buyer may have against Seller with respect to (i) a breach of any of Seller’s Warranties, (ii) a breach of any Surviving Covenants, or (iii) any acts constituting fraud by Seller.
\end{quote}

\subsection{Segmentation}
LexNLP's sentence model is a Punkt model trained using NLTK; however, the pretrained model that NLTK ships with does not perform well when presented with common legal abbreviations.
For example, NLTK and other NLP packages incorrectly parse Example 3, tripping up on the ``U.S.C'' abbreviation for the United States Code.

\begin{minted}[breaklines]{python}
>>> len(nltk.tokenize.PunktSentenceTokenizer()\
    .tokenize(text3))
5
>>> len(lexnlp.nlp.en.segments.sentences\
    .get_sentence_list(text3))
3
\end{minted}

\subsection{Extraction}
LexNLP allows for simple extraction of common information from documents such as our examples.  For example, parties, dates, definitions/terms, and geopolitical entities can be extracted from the contract preamble in Example 1:

\begin{minted}[breaklines]{python}
>>> from lexnlp.extract.en.entities.nltk_maxent import get_companies
>>> list(get_companies(text1))
[('WESLEY VILLAGE DEVELOPMENT', 'LP'),
 ('KBS-LEGACY APARTMENT COMMUNITY REIT VENTURE', 'LLC')]
\end{minted}

\begin{minted}[breaklines]{python}
>>> from lexnlp.extract.en.dates import get_dates
>>> list(get_dates(text1))
[datetime.date(2012, 10, 12)]
\end{minted}

\begin{minted}[breaklines]{python}
>>> from lexnlp.extract.en.definitions import get_definitions
>>> list(get_definitions(text1))
['Effective Date', 'Seller', 'Buyer']
\end{minted}

\begin{minted}[breaklines]{python}
>>> from lexnlp.extract.en.definitions import get_definitions
>>> list(get_definitions(text1))
['Effective Date', 'Seller', 'Buyer']
\end{minted}

Extracting geoentity references relies on the LexPredict Legal Dataset referenced above, and some configuration is required to select which set of geoentities and translations are desired.  Ambiguities are common when all translations and entities are loaded.  Examples are available in the test cases for LexNLP and in the ContraxSuite GitHub repository at \url{https://github.com/LexPredict/lexpredict-contraxsuite}. For example, if all translations of countries are loaded, then the German name for Iceland, \textit{Island}, can create frequent false positives in English text.  Researchers can handle issue like this by automatically identifying the most likely language for each unit of text and loading only relevant data; there are, however, many counterexamples of multilingual text that incorporate foreign names without translation, such as corporate addresses in contract preambles like Example 1.

Example 2 demonstrates a common style choice of legal documents, with amounts spelled out instead of being presented as numerals.  LexNLP features robust support for amounts written out in this fashion.  For example, ``one\-thousand fifty seven'',  ``one thousand fifty seven'', ``one thousand and fifty seven'', and ``one thousand fifty\-seven'' all parse to 1057 with LexNLP's \textsc{get\_amounts} method.  In Example 2, these monetary amounts parse as follows:

\begin{minted}[breaklines]{python}
>>> list(lexnlp.extract.en.money.get_money(text2))
[(1250000, 'USD'),
 (1250000.0, 'USD'),
 (250000, 'USD'),
 (250000.0, 'USD'),
 (1000000, 'USD'),
 (1000000.0, 'USD')]
\end{minted}

Example 2 also demonstrates the use of definitions or terms:
\begin{minted}[breaklines]{python}
>>> from lexnlp.extract.en.definitions import get_definitions
>>> list(get_definitions(text2))
['First Deposit', 'Deposit', 'Second Deposit']
\end{minted}

Finally, in addition to sentence boundary pitfalls, Example 3 also demonstrates the usage of constraints and regulatory references.

\begin{minted}[breaklines]{python}
>>> from lexnlp.extract.en.constraints import get_constraints
>>> constraints = list(get_constraints(text3))
>>> len(constraints)
5
>>> constraints[0]
('before',
 'by accepting the deed and closing the transaction, buyer, on behalf of itself and its successors and assigns, shall thereby release each of the seller parties from, and waive any and all liabilities against each of the seller parties for, attributable to, or in connection with the property, whether arising or accruing',
 '')
\end{minted}

Common regulations as provided in the repository at \url{https://github.com/LexPredict/lexpredict-legal-dictionary} are automatically identified, although some care must be taken to ensure that symbols such as \S \ are not transcoded improperly.
\begin{minted}[breaklines]{python}
>>> from lexnlp.extract.en.regulations import get_regulations
>>> list(get_regulations(text3))
[('United States Code', '42 USC 6901')]
\end{minted}

Citations to court opinions, state or federal codes, regulatory publications, or even international treaties are all important sources of information for many problems.  While tools for semantic analysis are becoming more valuable, legal text often conveys important semantic information in the form of citations that are otherwise absent from the text per se.  Once extracted document to document citations are useful metadata which can be interrogated using graph based methods.  (\cite{boulet2018network, bommarito2010distance})
  
\section{Acknowledgements}
\label{S:acknowledgments}
We would like to acknowledge the support of our company, LexPredict, and the developers and analysts who have helped in the design, development, maintenance, and testing of this software.  We would also like to acknowledge the incalculable contribution of the teams behind NLTK, numpy, scipy, scikit-learn, and gensim, as well as the Python team itself; without the ecosystem created by their work, this software would not exist.



\bibliographystyle{model1-num-names}
\bibliography{lexnlp.bib}







\end{document}